\documentclass{article}

 \usepackage[preprint]{neurips_2025}

\usepackage[utf8]{inputenc} 
\usepackage[T1]{fontenc}    
\usepackage{hyperref}       
\usepackage{url}            
\usepackage{booktabs}       
\usepackage{amsfonts}       
\usepackage{nicefrac}       
\usepackage{microtype}      
\usepackage[table]{xcolor}  
\usepackage{graphicx}
\usepackage{amsmath} 
\definecolor{latentgrpopurple}{RGB}{245,238,250}
\title{Latent-GRPO: Group Relative Policy Optimization for Latent Reasoning}

\author{%
  \normalfont\normalsize\mdseries
  \begin{tabular}[t]{@{}c@{\hspace{1.4em}}c@{\hspace{1.4em}}c@{\hspace{1.4em}}c@{}}
  Jingcheng Deng$^{1,2,*}$ & Zihao Wei$^{1,2,*}$ & Liang Pang$^{1,\dagger}$ & Junhong Wu$^{2,3}$
  \end{tabular}\\[0.35em]
  \normalfont\normalsize\mdseries
  \begin{tabular}[t]{@{}c@{\hspace{2.2em}}c@{\hspace{2.2em}}c@{}}
  Shicheng Xu$^{1,2}$ & Zenghao Duan$^{1,2}$ & Huawei Shen$^{1,2}$
  \end{tabular}\\[0.7em]
  \normalfont\small\mdseries
  $^1$State Key Laboratory of AI Safety, Institute of Computing Technology,\\
  Chinese Academy of Sciences \\
  $^2$University of Chinese Academy of Sciences \\
  $^3$Institute of Automation, Chinese Academy of Sciences \\
  \texttt{\{dengjingcheng23s, pangliang\}@ict.ac.cn} \\
  $^{*}$Equal contribution. \quad $^{\dagger}$Corresponding author.
}

\begin{document}

\maketitle

\begin{abstract}
  Latent reasoning offers a more efficient alternative to explicit reasoning by compressing intermediate reasoning into continuous representations and substantially shortening reasoning chains. However, existing latent reasoning methods mainly focus on supervised learning, and reinforcement learning in latent space remains highly unstable. We study this problem through the lens of Group Relative Policy Optimization (GRPO), and show that directly adapting GRPO to latent reasoning is fundamentally non-trivial: latent reasoning changes both the probability density and the sampling mechanism, causing three coupled bottlenecks: absence of intrinsic latent manifolds, where unconstrained exploration pushes rollouts off the valid latent manifold; exploration-optimization misalignment, where trajectory-level rewards can induce incorrect token-level updates; and latent mixture non-closure, where jointly reinforcing multiple correct latent paths can produce an invalid averaged state. To address them, we propose \textbf{Latent-GRPO}, which combines invalid-sample advantage masking, one-sided noise sampling, and optimal correct-path first-token selection. Across four low-difficulty benchmarks (e.g., GSM8K-Aug) and four high-difficulty benchmarks (e.g., AIME), Latent-GRPO improves over its latent initialization by 7.86 Pass@1 points on low-difficulty tasks and surpasses explicit GRPO by 4.27 points on high-difficulty tasks while using 3--4$\times$ shorter reasoning chains. It also achieves stronger pass@$k$ performance under Gumbel sampling. These results establish Latent-GRPO as an effective approach for stable and efficient latent reasoning. Our code, models, and data are available at \url{https://github.com/DJC-GO-SOLO/Latent-GRPO}.
\end{abstract}

\section{Introduction}
\label{intro}
Large Language Models (LLMs) have demonstrated remarkable efficacy in complex reasoning tasks through the generation of discrete chains of thought (CoT) ~\citep{DBLP:conf/nips/Wei0SBIXCLZ22, DBLP:conf/aaai/XuPZGWDPSC26}. While test-time scaling studies show that performance often improves with longer reasoning chains~\citep{DBLP:conf/emnlp/MuennighoffYSLFHZLCH25, DBLP:journals/corr/abs-2412-16720}, this paradigm incurs substantial computational redundancy and latency ~\citep{DBLP:journals/corr/abs-2508-17627}. To reduce this cost, recent work has explored \textit{latent reasoning}, which shifts intermediate reasoning from explicit language into continuous representations. Early methods ~\citep{DBLP:journals/corr/abs-2412-06769, DBLP:journals/corr/abs-2505-16552}, which directly adopts the hidden state as the latent token, often achieve shorter reasoning chains at the expense of degraded performance. Recently, vocabulary-superposition approaches ~\citep{DBLP:journals/corr/abs-2505-15778}, which adopts a linear combination of the top-$k$ vocabulary as the latent token, have emerged as an alternative. In particular, Latent-SFT ~\citep{DBLP:journals/corr/abs-2510-15522} demonstrates that with proper Supervised Fine-Tuning (SFT), latent reasoning can match or even surpass explicit supervised fine-tuning while substantially compressing the reasoning chain.

With the success of SFT approaches, Reinforcement Learning naturally emerges as the next key step. Previous works (see Section~\ref{sec:preliminaries} for details) have established the fundamental tools for RL training: (1) \textbf{Sampling Mechanism:} ~\citet{DBLP:journals/corr/abs-2508-03440} proposes to injects Gumbel noise to simulate the explicit sampling process, achieving effective exploration; (2) \textbf{Probability Density Estimation:} Soft-GRPO ~\citep{DBLP:journals/corr/abs-2511-06411} adopts the Gumbel re-parameter trick to estimate the probability density of the latent token, enabling PPO-style training. However, even with these established techniques, RL for latent reasoning remains an unsolved problem. Soft-GRPO struggles to outperform standard token RL, and scaling the Gumbel temperature to encourage broader exploration invariably precipitates catastrophic model collapse. Overall, an effective RL framework to fully unleash the potential of latent reasoning remains a gap.

\begin{figure}[!t]
  \centering
  \includegraphics[width=1.\linewidth]{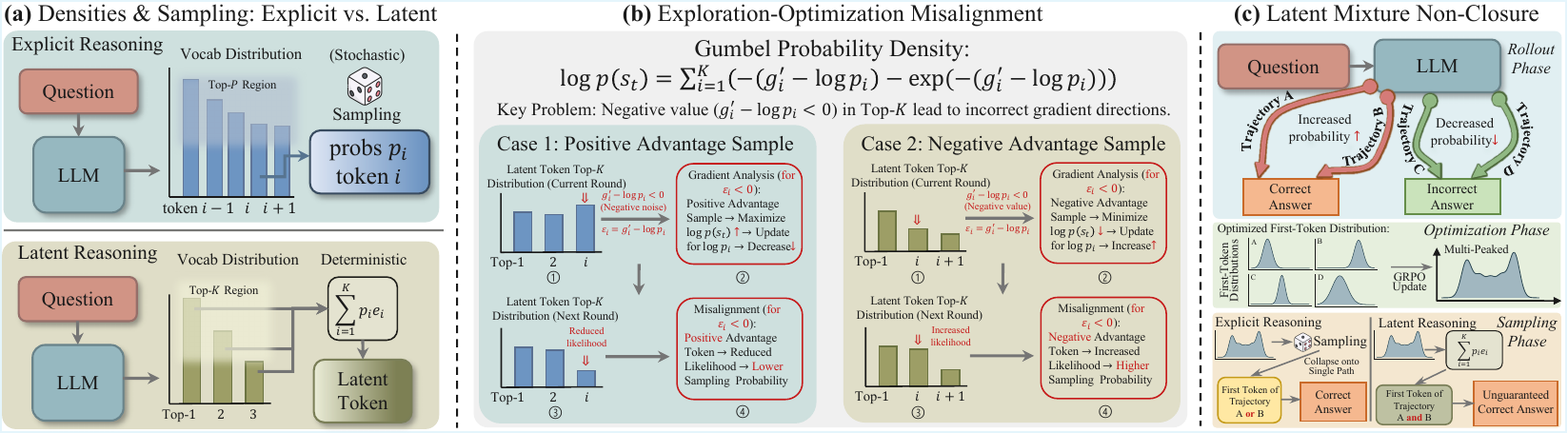}
  \caption{Key discrepancies and bottlenecks in adapting GRPO to latent-token reasoning. (a) Explicit and latent reasoning differ in both probability density and sampling mechanism. (b) Exploration-Optimization Misalignment: The standard Gumbel density objective yields incorrect directional updates, inadvertently pushing down components from positive-advantage latent trajectories while pushing up components from negative-advantage ones. (c) Latent Mixture Non-Closure: jointly reinforcing several correct latent paths can produce an averaged continuous state that does not correspond to a valid reasoning path.}
  \label{fig_overview}
\end{figure}

In this paper, we study this problem with the most widely adopted RL approach for LLMs--GRPO~\citep{DBLP:journals/corr/abs-2402-03300}. We identified three fundamental bottlenecks: (1) \textbf{Absence of Intrinsic Latent Manifolds}: direct application of Latent-RL fails to spontaneously induce latent reasoning capabilities within explicit models; such foundational structures must be strictly injected prior to RL exploration, typically during the SFT phase. Furthermore, unbounded continuous exploration can easily propel the model into chaotic, out-of-distribution (OOD) regions off the valid manifold, resulting in non-terminating generation. (2) \textbf{Exploration-Optimization Misalignment:} under naive Gumbel perturbations, the optimization direction can become mismatched with the trajectory-level advantage. As a result, components from positive-advantage latent trajectories may receive downward probability updates, while components from negative-advantage trajectories may receive upward updates (See Figure~\ref{fig_overview}(b)). (3) \textbf{Latent Mixture Non-Closure}: The set of valid latent reasoning states is not closed under jointly fitting multiple correct paths. In explicit reasoning, even when the model assigns high probabilities to several correct first-step continuations, discrete sampling ensures that reasoning proceeds along a single concrete path. In latent reasoning, however, the model must continue from a single continuous latent action under the same prefix, so jointly reinforced correct paths can induce an averaged latent state that need not support any correct continuation (See Figure~\ref{fig_overview}(c)).

Building on this diagnosis, we propose three tailored designs. First, we initialize the model from Latent-SFT and apply \textbf{Invalid Sample Advantage Masking} to discard overlength trajectories that likely fall off the valid latent reasoning manifold. Second, we introduce \textbf{One-sided Noise Sampling}, which constrains Gumbel perturbations to be strictly positive and thus aligns the update direction with the advantage sign. Third, we propose \textbf{Optimal Correct Path First Token Selection}, which scores each correct trajectory by its trajectory-level average surrogate log-probability under the rollout objective and updates only the first token of the selected correct path, while keeping later-token updates from other paths active. This specifically targets the shared initial step, where multiple correct trajectories under the same prefix are most vulnerable to harmful mode averaging. Together, these designs yield \textbf{Latent-GRPO}, a stable RL framework for latent-token reasoning.

Extensive experiments on LLaMA-3.2-1B-Instruct and Qwen2.5-Math-7B show that Latent-GRPO delivers stable gains across both low- and high-difficulty settings. On low-difficulty tasks, it improves over Latent-SFT by 7.86 Pass@1 points on average while using a 4.44$\times$ shorter reasoning chain than explicit GRPO. On high-difficulty tasks, the advantage becomes even clearer: it improves over Latent-SFT by 14.77 points on average, outperforms explicit GRPO by more than 4 Pass@1 points, uses a 3.31$\times$ shorter reasoning chain, and achieves the best Pass@1 on Math500, AIME24, and AIME25. Under Gumbel sampling, Latent-GRPO also retains strong pass@$k$ capability, reaching 50+ pass@$64$ on the AIME benchmarks. These results validate Latent-GRPO as an effective and robust framework for efficient, high-performance latent reasoning.

\section{Preliminaries}
\label{sec:preliminaries}
\subsection{Latent Reasoning}
\label{sec:latent_reasoning}
Standard LLMs generate explicit Chains of Thought (CoT) via an autoregressive process, where each discrete token $y_t$ is sampled from the vocabulary distribution $P(y_t | \boldsymbol{x}, y_{<t})$. To alleviate the computational redundancy of discrete sequences, latent reasoning replaces intermediate explicit tokens with continuous vector representations, termed \textit{latent tokens}. 

At generation step $t$, instead of sampling a discrete token, the model computes a probability distribution over the vocabulary. To maintain computational tractability and filter long-tail noise, this distribution is typically truncated to the top-$K$ candidates. Let $\mathcal{V}_t^{(K)}$ denote the set of the top-$K$ tokens at step $t$, and let $p_{t,i}$ denote the normalized probability of the $i$-th token in $\mathcal{V}_t^{(K)}$. The latent token $s_t$ is defined as the expectation of the corresponding top-$K$ embeddings~\citep{DBLP:journals/corr/abs-2505-15778, DBLP:journals/corr/abs-2505-14827}:
\begin{equation}
    s_t = \sum_{i=1}^K p_{t,i} \boldsymbol{e}_{t,i},
\end{equation}
where $\boldsymbol{e}_{t,i} \in \mathbb{R}^d$ is the embedding vector of the $i$-th token in $\mathcal{V}_t^{(K)}$. This construction maps a truncated vocabulary distribution into a continuous latent action in the embedding space.


\subsection{Reinforcement Learning via Soft-GRPO}
To further optimize reasoning capabilities without relying on extensive supervised data, Group Relative Policy Optimization (GRPO)~\citep{DBLP:journals/corr/abs-2402-03300} is widely adopted. For a given query $\boldsymbol{x}$, GRPO samples a group of $G$ outputs $\{o_1, o_2, \dots, o_G\}$, computes their rule-based rewards $\{R_1, R_2, \dots, R_G\}$, and calculates the normalized advantage $\hat{A}_j = (R_j - \mu_R)/\sigma_R$ for each output.

Adapting GRPO to the latent-token paradigm requires redefining the probability density to account for a mixed generation trajectory: a continuous latent reasoning segment $s_{1:T_{\text{lat}}}$ followed by a discrete explicit answer segment $y_{1:T_{\text{exp}}}$. \textbf{Soft-GRPO}~\citep{DBLP:journals/corr/abs-2511-06411} models these two phases separately.

For the latent segment ($t \le T_{\text{lat}}$), Soft-GRPO injects Gumbel noise $\xi_i$ into the top-$K$ components to simulate the stochasticity of explicit sampling and encourage exploration. Let $g'_{t,i} = \log p_{t,i} + \xi_i$ denote the Gumbel-perturbed score of the $i$-th component. During rollout, these perturbed scores are converted into noisy top-$K$ mixture weights through a temperature-scaled softmax:
\begin{equation}
\alpha_{t,i}
=
\frac{\exp(g'_{t,i}/\tau_g)}
{\sum_{u=1}^{K}\exp(g'_{t,u}/\tau_g)},
\qquad
s_t = \sum_{i=1}^{K}\alpha_{t,i}\boldsymbol{e}_{t,i},
\label{eq:noisy_latent_token}
\end{equation}
where $\tau_g$ is the Gumbel temperature. Thus, the injected noise affects the forward latent token through the auxiliary perturbed-score representation. Based on the same perturbations, Soft-GRPO defines the density of each latent token, which we refer to as the \textit{Gumbel-perturbation density}, as:
\begin{equation}
\log \pi_\theta(s_t) = \sum_{i=1}^K \left( -(g'_{t,i} - \log p_{t,i}) - \exp(-(g'_{t,i} - \log p_{t,i})) \right).
\end{equation}

For the explicit segment ($t > T_{\text{lat}}$), the model reverts to standard autoregressive generation. The probability of generating the discrete token $y_t$ is given by the standard categorical distribution over the vocabulary $\mathcal{V}$:
\begin{equation}
\log \pi_\theta(y_t) = \log P(y_t \mid \boldsymbol{x}, s_{1:T_{\text{lat}}}, y_{<t}).
\end{equation}

Consequently, the policy ratio $r_{j,t}(\theta)$ utilized in the proximal policy update is piecewise defined, seamlessly bridging the continuous and discrete spaces:

\begin{equation}
r_{j,t}(\theta)=
\begin{cases}
\dfrac{\pi_\theta\!\left(s_{j,t}\mid \boldsymbol{x}, s_{j,<t}\right)}
{\pi_{\mathrm{old}}\!\left(s_{j,t}\mid \boldsymbol{x}, s_{j,<t}\right)},
& t \le T_{\mathrm{lat}}, \\[1.2ex]
\dfrac{\pi_\theta\!\left(y_{j,t}\mid \boldsymbol{x}, s_{j,1:T_{\mathrm{lat}}}, y_{j,<t}\right)}
{\pi_{\mathrm{old}}\!\left(y_{j,t}\mid \boldsymbol{x}, s_{j,1:T_{\mathrm{lat}}}, y_{j,<t}\right)},
& t > T_{\mathrm{lat}}.
\end{cases}
\end{equation}

Based on the Gumbel-perturbation density for latent tokens and the standard categorical density for explicit tokens, Soft-GRPO optimizes a clipped surrogate objective with a step-wise Kullback--Leibler (KL) penalty to constrain policy drift. Let
\begin{equation}
L_j = T_{\mathrm{lat}}^{(j)} + T_{\mathrm{exp}}^{(j)}
\end{equation}
denote the total length of the $j$-th trajectory, and let
\begin{equation}
\mathbb{D}_{\mathrm{KL}}^{(j,t)}
=
\mathbb{D}_{\mathrm{KL}}
\!\left(
\pi_\theta(\cdot \mid h_{j,t})
\,\|\, 
\pi_{\mathrm{ref}}(\cdot \mid h_{j,t})
\right)
\end{equation}
denote the KL divergence at step $t$ under the conditioning context $h_{j,t}$. The Soft-GRPO objective is then given by
\begin{equation}
\begin{aligned}
\mathcal{J}(\theta)
&=
\mathbb{E}_{\boldsymbol{x}\sim P(\mathcal{X})}
\Bigg[
\frac{1}{G}
\sum_{j=1}^{G}
\frac{1}{L_j}
\sum_{t=1}^{L_j}
\Bigg(
\min\!\Big(
r_{j,t}(\theta)\hat{A}_j, \\
&\qquad\qquad\qquad
\operatorname{clip}\!\big(r_{j,t}(\theta),\,1-\epsilon_{\mathrm{clip}},\,1+\epsilon_{\mathrm{clip}}\big)\hat{A}_j
\Big)
-
\beta\,\mathbb{D}_{\mathrm{KL}}^{(j,t)}
\Bigg)
\Bigg].
\end{aligned}
\end{equation}
where $\epsilon_{\mathrm{clip}}$ denotes the PPO clipping parameter.

This objective has a gradient-level source of Exploration-Optimization Misalignment. For the standard two-sided Gumbel density, the local update direction of a latent component is proportional to $\hat{A}_j\left(1-\exp(-\Delta_i)\right)$, where $\Delta_i=g'_i-\log p_i(\theta)$ is the perturbation margin. Thus, even when $\hat{A}_j>0$, a negative margin $\Delta_i<0$ pushes the corresponding component probability downward. Conversely, for a negative-advantage trajectory, a negative margin may push the component probability upward. A detailed derivation is provided in Appendix~\ref{app:gradient_misalignment}.

While Soft-GRPO provides a mathematical bridge from GRPO to mixed latent-discrete trajectories, its unconstrained Gumbel-based exploration introduces substantial stability challenges in latent reasoning, which motivate the design of our method.

\section{Methodology: Latent-GRPO}
To systematically overcome the fundamental bottlenecks of continuous latent reasoning---namely the absence of intrinsic manifolds, exploration-optimization misalignment, and latent mixture non-closure---we propose \textbf{Latent-GRPO}. Building upon the mathematical foundation of Soft-GRPO, our framework introduces three carefully designed strategies: (1) Invalid Sample Advantage Masking to establish exploration guardrails, (2) One-sided Noise Sampling to correct optimization directions, and (3) Optimal Correct Path First Token Selection to avoid harmful mode averaging.

\subsection{Invalid Sample Advantage Masking for Bounded Exploration}
As observed in Soft-GRPO, latent RL alone cannot reliably induce latent reasoning capabilities in an explicit model from scratch. Latent-GRPO therefore starts from a Latent-SFT-initialized policy to provide a valid latent reasoning manifold before RL exploration. However, even with such initialization, unconstrained Gumbel exploration can still push rollout trajectories off this manifold, often causing chaotic, non-terminating generation. If these invalid trajectories are included in standard GRPO advantage normalization, they contaminate the group statistics and corrupt the normalized advantages of valid samples. To prevent this, we introduce \textbf{Invalid Sample Advantage Masking} before group-wise normalization.

For a group of $G$ sampled trajectories, let $L_j$ denote the total length of the $j$-th trajectory, and let $L_{\max}$ be the predefined maximum response length. Since rollouts are truncated at $L_{\max}$, we treat a trajectory as valid only if it terminates before reaching this limit, i.e., it emits an EOS/stop token within the allowed budget. We define the subset of valid trajectories as
\begin{equation}
\mathcal{G}_{\mathrm{valid}} = \{\, j \mid L_j < L_{\max} \,\}.
\end{equation}
The group baseline is computed only over this valid subset:
\begin{equation}
\mu_R = \frac{1}{|\mathcal{G}_{\mathrm{valid}}|}\sum_{j\in\mathcal{G}_{\mathrm{valid}}} R_j,
\qquad
\sigma_R =
\sqrt{
\frac{1}{|\mathcal{G}_{\mathrm{valid}}|}
\sum_{j\in\mathcal{G}_{\mathrm{valid}}}(R_j-\mu_R)^2
}.
\end{equation}
The valid-sample normalized advantage is then defined as
\begin{equation}
\hat{A}_j^{\mathrm{valid}} =
\begin{cases}
\dfrac{R_j-\mu_R}{\sigma_R},
& \text{if } j\in\mathcal{G}_{\mathrm{valid}}, \\[1.2ex]
0,
& \text{if } j\notin\mathcal{G}_{\mathrm{valid}}.
\end{cases}
\end{equation}
For numerical stability, if $|\mathcal{G}_{\mathrm{valid}}|=0$ or $\sigma_R=0$, we set $\hat{A}_j^{\mathrm{valid}}=0$ for all $j$. By excluding trajectories that fail to terminate before the maximum response length from the group statistics and explicitly assigning them zero advantage, Invalid Sample Advantage Masking prevents off-manifold rollouts from interfering with the optimization of valid trajectories, thereby keeping exploration within the effective latent reasoning region.

\subsection{One-sided Noise Sampling for Aligned Optimization}
\label{sec:one_sided_noise_sampling}
The standard Gumbel-perturbation density in Soft-GRPO injects unconstrained noise $\xi_i$, yielding the perturbed target $g'_i = \log p_i + \xi_i$. Since $\xi_i$ can be negative, $g'_i$ may fall below the original prediction $\log p_i$. This creates \textit{Exploration-Optimization Misalignment} described in Section~\ref{intro}: for positive-advantage trajectories, the probability mass of the sampled latent components may be reduced when it should be increased, while for negative-advantage trajectories, it may be increased when it should be reduced. 

To correct this optimization-direction mismatch, we propose \textbf{One-sided Noise Sampling}, which constructs a strictly positive perturbation margin for each latent component. Specifically, we first sample i.i.d. standard Gumbel noise and then apply a per-dimension clipped-and-shifted transformation:
\begin{equation}
\xi_i^+ = \operatorname{clip}(\xi_i, -a, b) + a + \delta,
\qquad
\xi_i \overset{\mathrm{i.i.d.}}{\sim} \mathrm{Gumbel}(0,1),
\end{equation}
where $a>0$ and $b>0$ are the lower and upper clipping magnitudes, respectively, and $\delta>0$ is a small positive offset. In our implementation, we use $a=1.5$ and $b=3.0$, corresponding to the common Gumbel clipping range $[-1.5, 3.0]$. We use this transformation as a target-margin construction rather than requiring the transformed perturbation to remain Gumbel-distributed. Since the transformation is applied independently to each dimension, it avoids coupling the top-$K$ components while guaranteeing $\xi_i^+ \in [\delta, a+b+\delta]$ for all $i$. Moreover, the uniform shift $a+\delta$ does not change the normalized top-$K$ weights because softmax is invariant to adding the same constant to all perturbed scores; it only moves the perturbation margins into the positive range. The resulting one-sided target is then written as
\begin{equation}
g_i^* = \log p_i(\theta_{\mathrm{old}}) + \xi_i^+,
\end{equation}
which satisfies $g_i^* > \log p_i(\theta_{\mathrm{old}})$ at the beginning of optimization. The rollout latent token is then constructed by the same temperature-scaled softmax in Eq.~\eqref{eq:noisy_latent_token}, with $g_i^*$ replacing $g'_i$. However, PPO-style training performs multiple update epochs on the same rollout data, so the updated prediction $\log p_i(\theta)$ may eventually exceed this fixed target. Once the target is crossed, the original one-sided assumption is violated because the effective perturbation margin becomes negative:
\begin{equation}
\Delta_i(\theta) = g_i^* - \log p_i(\theta).
\end{equation}
When $\Delta_i(\theta) < 0$, optimization is no longer guided by a positive perturbation margin. To preserve the intended one-sided behavior throughout repeated PPO updates, we apply a conditional Straight-Through Estimator (STE) directly to this margin. Specifically, we define the one-sided margin as
\begin{equation}
\widetilde{\Delta}_i(\theta)=
\begin{cases}
\mathrm{FlipGrad}\!\left(\Delta_i(\theta)\right),
& \text{if } \Delta_i(\theta) < 0, \\[1ex]
\Delta_i(\theta),
& \text{otherwise},
\end{cases}
\end{equation}
where $\mathrm{FlipGrad}(x)$ acts as the identity in the forward pass but negates the backward gradient, i.e.,
\begin{equation}
\nabla_x \mathrm{FlipGrad}(x) = -1.
\end{equation}
In this way, the PPO ratio still uses the original forward value, while the backward pass treats a crossed target as if its perturbation margin were reflected back to the positive side. Using this one-sided margin, we define the one-sided Gumbel-margin surrogate log-likelihood as
\begin{equation}
\log \pi_\theta^{+}(s_t)
=
\sum_{i=1}^{K}
\left(
-\widetilde{\Delta}_i(\theta)
-
\exp\!\left(-\widetilde{\Delta}_i(\theta)\right)
\right).
\end{equation}
Because the conditional STE modifies the backward pass, $\pi_\theta^{+}$ should be interpreted as a surrogate latent likelihood used for policy optimization rather than a mathematically exact probability density. In particular, this surrogate optimizes the auxiliary perturbed-score representation associated with a latent token, rather than a representation-invariant density over the embedding $s_t$ alone. Once the fixed one-sided target is exceeded, the gradient is flipped so that optimization continues to follow the intended one-sided direction throughout repeated policy updates.

\subsection{Optimal Correct Path First Token Selection for Avoiding Path Harmful Averaging}
\label{sec:first_token_selection}
Continuous latent reasoning faces a deterministic continuous-action bottleneck. Although Latent-GRPO optimizes a PPO-style objective, the latent token fed into the next step is a single continuous state rather than a sampled discrete branch. This creates a unique geometric pitfall termed \textit{Latent Mixture Non-Closure}. In standard GRPO, multiple correct trajectories in the same group can simultaneously receive positive training signals. This is usually harmless for explicit reasoning because categorical sampling ultimately commits generation to one concrete token and hence one concrete path. In latent reasoning, however, jointly reinforcing several correct latent paths under the same prefix can instead induce mode averaging: the fitted latent token may move toward a barycentric state between several correct paths, and this averaged state need not correspond to any valid continuation. This is analogous to the multi-modal action-averaging problem in deterministic continuous control, where fitting several valid actions with one continuous action can yield an invalid action between modes.

This problem is most severe at the first reasoning step ($t=1$). Since the prompt context $\boldsymbol{x}$ is identical across all trajectories, positive training signals from multiple correct trajectories are applied to the same prefix-to-action mapping. Therefore, jointly fitting multiple correct first-step paths can push the first latent token toward a harmful averaged state under the same prefix. For later steps ($t>1$), the generated prefixes have already diverged, which substantially alleviates this mode-averaging effect.

To address this issue, we propose \textbf{Optimal Correct Path First Token Selection}. Let $\mathcal{G}_{\text{correct}} \subseteq \mathcal{G}$ denote the subset of trajectories in the group whose final answers are verified as correct by the task-specific reward function or verifier. When $|\mathcal{G}_{\text{correct}}| > 1$, we compute a trajectory-level score for each correct path using its average per-token surrogate log-probability under the rollout objective:
\begin{equation}
\bar{\ell}_j
=
\frac{1}{L_j}
\left[
\sum_{t=1}^{T_{\mathrm{lat}}^{(j)}}
\log \pi_{\theta_{\mathrm{old}}}^{+}\!\left(s_{j,t}\mid \boldsymbol{x}, s_{j,<t}\right)
+
\sum_{t=T_{\mathrm{lat}}^{(j)}+1}^{L_j}
\log \pi_{\theta_{\mathrm{old}}}\!\left(y_{j,t}\mid \boldsymbol{x}, s_{j,1:T_{\mathrm{lat}}^{(j)}}, y_{j,<t}\right)
\right].
\end{equation}
We then select the optimal trajectory as
\begin{equation}
j^* = \arg\max_{j \in \mathcal{G}_{\text{correct}}} \bar{\ell}_j.
\end{equation}
This criterion selects the correct path with the highest average surrogate log-probability under the rollout objective, providing a conservative tie-breaking rule among multiple verified correct trajectories without inducing harmful mode averaging at the shared initial step. Importantly, this path selection is used only to decide which first latent token remains active at $t=1$; later-token updates from the other paths are preserved once their prefixes diverge. We then define a time-dependent and trajectory-dependent mask
\begin{equation}
M_{j,t} =
\begin{cases}
0, & \text{if } j \in \mathcal{G}_{\text{correct}} \setminus \{j^*\} \text{ and } t = 1, \\[1ex]
1, & \text{otherwise},
\end{cases}
\end{equation}
and use it to construct the masked advantage
\begin{equation}
\tilde{A}_{j,t} = M_{j,t}\hat{A}_j^{\mathrm{valid}}.
\end{equation}
As a result, only the first latent token of the selected optimal correct path contributes to the update at $t=1$, while all competing correct paths remain active at subsequent steps. Thus, the method does not collapse optimization to a single full trajectory; it only avoids harmful mode averaging at the shared initial state without suppressing later-stage latent exploration.

\subsection{Final Latent-GRPO Objective}
Combining the three strategies above, the final Latent-GRPO objective is defined as
\begin{equation}
\begin{aligned}
\mathcal{J}_{\mathrm{Latent\text{-}GRPO}}(\theta)
&=
\mathbb{E}_{\boldsymbol{x}\sim P(\mathcal{X})}
\Bigg[
\frac{1}{G}
\sum_{j=1}^{G}
\frac{1}{L_j}
\sum_{t=1}^{L_j}
\Bigg(
\min\!\Big(
r_{j,t}^{*}(\theta)\,\tilde{A}_{j,t}, \\
&\qquad\qquad\qquad
\operatorname{clip}\!\big(
r_{j,t}^{*}(\theta),\,1-\epsilon_{\mathrm{clip}},\,1+\epsilon_{\mathrm{clip}}
\big)\tilde{A}_{j,t}
\Big)
-\beta\,\mathbb{D}_{\mathrm{KL}}^{(j,t)}
\Bigg)
\Bigg].
\end{aligned}
\end{equation}
Here, $\epsilon_{\mathrm{clip}}$ denotes the PPO clipping parameter, $\tilde{A}_{j,t}$ combines Invalid Sample Advantage Masking and Optimal Correct Path First Token Selection, and the modified PPO-style surrogate ratio is defined as
\begin{equation}
r_{j,t}^{*}(\theta)=
\begin{cases}
\dfrac{\pi_\theta^{+}\!\left(s_{j,t}\mid \boldsymbol{x}, s_{j,<t}\right)}
{\pi_{\theta_{\mathrm{old}}}^{+}\!\left(s_{j,t}\mid \boldsymbol{x}, s_{j,<t}\right)},
& t \le T_{\mathrm{lat}}^{(j)}, \\[1.2ex]
\dfrac{\pi_\theta\!\left(y_{j,t}\mid \boldsymbol{x}, s_{j,1:T_{\mathrm{lat}}^{(j)}}, y_{j,<t}\right)}
{\pi_{\theta_{\mathrm{old}}}\!\left(y_{j,t}\mid \boldsymbol{x}, s_{j,1:T_{\mathrm{lat}}^{(j)}}, y_{j,<t}\right)},
& t > T_{\mathrm{lat}}^{(j)}.
\end{cases}
\end{equation}

\section{Experiments}
\subsection{Experimental Setup}
We follow the experimental setting of Latent-SFT and divide our experiments into two regimes, low difficulty and high difficulty, so that each setting is paired with an appropriate latent-SFT initialization. Because the full training corpora of existing reasoning models are unavailable, we cannot construct equally sufficient latent-SFT initializations for them, making the comparison unfair. We therefore start from base instruct models and, within each regime, train both the explicit SFT initialization and the Latent-SFT initialization on matched data before comparing their RL performance.

\paragraph{Low-difficulty Tasks} We evaluate on GSM8K-Aug ~\citep{DBLP:journals/corr/abs-2405-14838}, GSM-Hard ~\citep{DBLP:conf/icml/GaoMZ00YCN23}, SVAMP ~\citep{DBLP:conf/naacl/PatelBG21}, and MultiArith ~\citep{DBLP:conf/emnlp/RoyR15}. The corresponding base model is LLaMA-3.2-1B-Instruct ~\citep{DBLP:journals/corr/abs-2407-21783}. Both the explicit SFT model and the Latent-SFT model are trained on the GSM8K-Aug training split. Because in-domain RL data are unavailable, the RL stage uses an LLM-generated dataset derived from the GSM8K-Aug training set.

\paragraph{High-difficulty Tasks} We evaluate on Math500 ~\citep{DBLP:conf/nips/HendrycksBKABTS21}, AIME24, AIME25 ~\citep{maa2024aime}, and GPQA ~\citep{DBLP:journals/corr/abs-2311-12022}. The corresponding base model is Qwen2.5-MATH-7B ~\citep{DBLP:journals/corr/abs-2409-12122}. Both the explicit SFT model and the Latent-SFT model are trained on the same subset of Open-R1 ~\citep{openr1} with 4K-length reasoning chains, and the RL stage uses DAPO-Math-En ~\citep{DBLP:journals/corr/abs-2503-14476}.

Additional implementation details are provided in Appendix~\ref{app:implementation_details}.

\subsection{Main Results}
\newcommand{\tbdres}{-}

\paragraph{Overall Comparison.} We organize the main results around two complementary comparisons. First, we compare \textit{explicit reasoning} and \textit{latent reasoning} under matched base models and data, in order to evaluate whether latent reasoning can preserve task performance while reducing reasoning length. Second, within the latent-reasoning setting, we further compare \textbf{Latent-GRPO} against \textbf{Soft-GRPO} to demonstrate the effectiveness of Latent-GRPO for latent reasoning; directly applying standard GRPO to train the Latent-SFT model consistently leads to training collapse, so we do not report its performance in the main tables. Across both settings, we report Pass@1 together with reasoning length (\#L), so that the results jointly reflect reasoning effectiveness and efficiency.

\begin{table*}[t]
\centering
\small
\setlength{\tabcolsep}{4.2pt}
\renewcommand{\arraystretch}{1.12}
\resizebox{\textwidth}{!}{%
\begin{tabular}{lcccccccccccc}
\toprule
& \multicolumn{2}{c}{GSM8K-Aug}
& \multicolumn{2}{c}{GSM-Hard}
& \multicolumn{2}{c}{SVAMP}
& \multicolumn{2}{c}{MultiArith}
& \multicolumn{4}{c}{Average} \\
\cmidrule(lr){2-3}\cmidrule(lr){4-5}\cmidrule(lr){6-7}\cmidrule(lr){8-9}\cmidrule(lr){10-13}
Method
& Pass@1 $\uparrow$ & \#L $\downarrow$
& Pass@1 $\uparrow$ & \#L $\downarrow$
& Pass@1 $\uparrow$ & \#L $\downarrow$
& Pass@1 $\uparrow$ & \#L $\downarrow$
& Pass@1 $\uparrow$ & \#L $\downarrow$ & $\Delta$ Pass@1 $\uparrow$ & GRPO/\#L $\uparrow$ \\
\midrule
\multicolumn{13}{c}{\textbf{LLaMA-3.2-1B-Instruct}} \\
\cmidrule(lr){1-13}
\multicolumn{13}{c}{\textit{Explicit Reasoning}} \\
SFT
& 49.51    & 77.46  & 11.67   & 85.78    & 57.38  & 44.10  & 92.23     & 45.49   & 52.70     & 63.20 & \tbdres & \tbdres \\
GRPO
& 62.26 & 111.1 & 14.70 & 130.8 & \textbf{60.78} & 69.40 & 94.16 & 65.50 & 57.98 & 94.20 & +5.28 & 1.00$\times$ \\
\cmidrule(lr){1-13}
\multicolumn{13}{c}{\textit{Latent Reasoning (No Sampling)}} \\
Latent-SFT
& 47.55 & 22.00 & 10.25 & 25.50 & 53.22 & 15.30 & 90.80 & 16.30 & 50.46 & 19.78 & \tbdres & \tbdres \\
Soft-GRPO
& 48.34 & 21.60 & 10.30 & 25.60 & 53.19 & 15.30 & 91.07 & 16.40 & 50.73 & 19.73 & +0.27 & 4.77$\times$ \\
\rowcolor{latentgrpopurple}
Latent-GRPO
& \textbf{66.29} & 23.80 & \textbf{15.83} & 28.00 & 56.50 & 15.80 & \textbf{94.65} & 17.20 & \textbf{58.32} & 21.20 & \textbf{+7.86} & 4.44$\times$ \\
\bottomrule
\end{tabular}
}
\caption{Main results on the low-difficulty benchmark suite. We report Pass@1 and reasoning length (\#L) for each dataset, together with averaged performance across benchmarks. Explicit reasoning uses the standard discrete-token CoT pattern, while latent reasoning uses the No Sampling mode. $\Delta$ denotes the average Pass@1 gain over the corresponding SFT baseline, i.e., GRPO over SFT for explicit reasoning and Soft-GRPO or Latent-GRPO over Latent-SFT for latent reasoning. GRPO/\#L denotes the average-length compression ratio relative to explicit GRPO, computed as the average reasoning length of explicit GRPO divided by that of the current RL method; larger values indicate shorter reasoning. ``-'' indicates that the corresponding quantity is not applicable.}
\label{tab:low_difficulty_main_results}
\end{table*}

\begin{figure}[t]
\centering
\includegraphics[width=\linewidth]{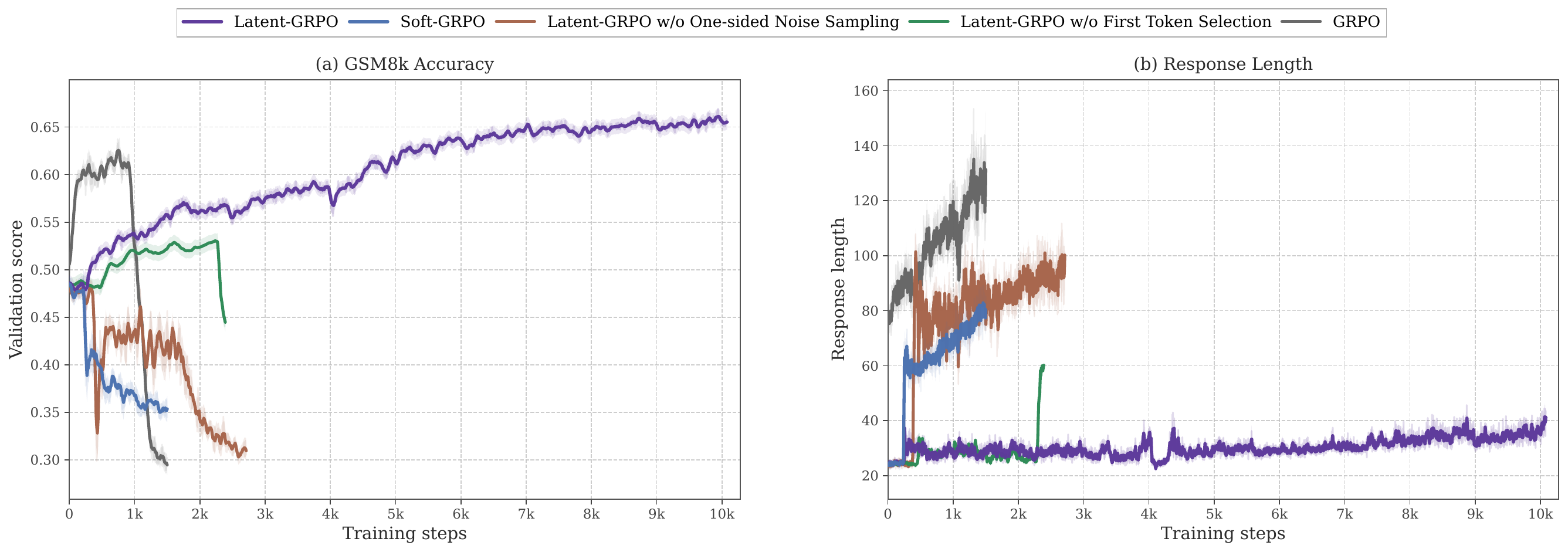}
\caption{RL training dynamics on the low-difficulty setting for LLaMA-3.2-1B-Instruct. The figure shows validation-set performance together with the evolution of response length during training on GSM8K-Aug.}
\label{fig:gsm8k_experiment_comparison}
\end{figure}

\paragraph{Low-difficulty Performance.} Table~\ref{tab:low_difficulty_main_results} reports the low-difficulty results for LLaMA-3.2-1B-Instruct. On the in-domain GSM8K-Aug benchmark, Latent-GRPO delivers a substantial improvement over the latent initialization, raising Pass@1 from 47.55 to 66.29, while the response length changes only slightly from 22.00 to 23.80. This indicates that the gain comes from better reasoning rather than simply generating longer responses. It also clearly outperforms Soft-GRPO (48.34) and even surpasses GRPO (62.26), while remaining much shorter than GRPO in response length (23.80 vs.\ 111.1). On the OOD benchmarks, the same trend largely persists: Latent-GRPO consistently improves over the latent baselines and achieves the best performance on GSM-Hard and MultiArith, suggesting that its gains are not tied to a single training distribution. Moreover, it maintains short responses on these OOD datasets, e.g., 28.00 vs.\ 130.8 on GSM-Hard and 17.20 vs.\ 65.50 on MultiArith when compared with GRPO. Although Latent-GRPO is slightly below GRPO on SVAMP (56.50 vs.\ 60.78), we attribute this gap to the mismatch between the explicit and latent starting points on that dataset, since the SFT and Latent-SFT models are not equally strong there. Consistent with this interpretation, the average improvement of Latent-GRPO over Latent-SFT is 7.86 points, exceeding the 5.28-point gain of GRPO over SFT. At the same time, it remains 4.44$\times$ shorter than GRPO on average. Figure~\ref{fig:gsm8k_experiment_comparison} provides a complementary view of the training dynamics. In the early stage, Latent-GRPO converges more slowly than GRPO, which is expected because, beyond optimizing token probabilities, it must jointly optimize both the top-$K$ token choices underlying each latent token and their relative mixture weights. As training proceeds, however, GRPO quickly collapses, whereas Latent-GRPO remains stable throughout the full 10k-step run. The response-length curves further show that Latent-GRPO exhibits almost no noticeable length growth during training.

\begin{figure}[t]
\centering
\includegraphics[width=\linewidth]{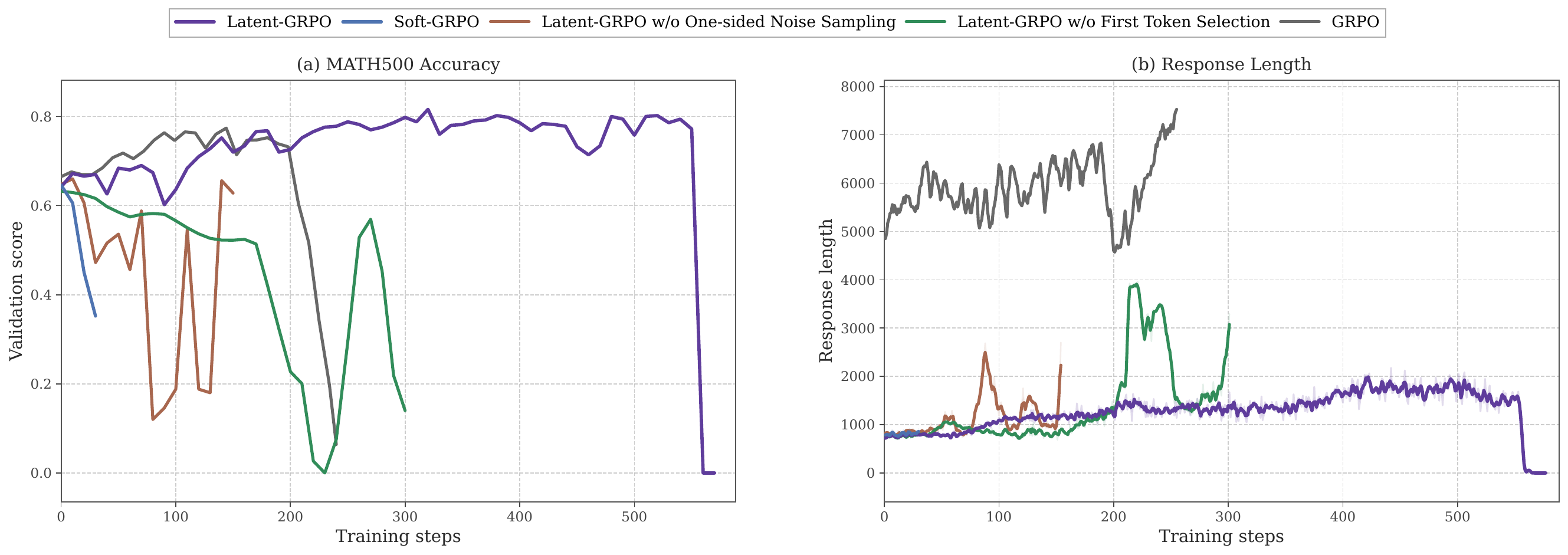}
\caption{RL training dynamics on the high-difficulty setting for Qwen2.5-Math-7B. The figure shows validation-set performance on Math500 together with the evolution of response length during training on DAPO-Math-En.}
\label{fig:math500_experiment_comparison}
\end{figure}

\paragraph{High-difficulty Performance.} Table~\ref{tab:high_difficulty_main_results} reports the results on the high-difficulty benchmark suite, where we evaluate the same comparison protocol on Qwen2.5-Math-7B across more challenging mathematical and scientific reasoning tasks. As task difficulty increases, the benefit of Latent-GRPO becomes even more pronounced. Compared with the low-difficulty setting, where Latent-GRPO improves over Latent-SFT by 7.86 points on average, the gain here rises to 14.77 points, nearly doubling the relative improvement. This advantage is also substantially larger than the 7.54-point gain achieved by explicit GRPO over SFT. Moreover, Latent-GRPO surpasses explicit GRPO by 4.27 points in average Pass@1, while still remaining 3.31$\times$ shorter in reasoning length on average. Consistent with this overall trend, it achieves the best Pass@1 on all three mathematical reasoning benchmarks: Math500, AIME24, and AIME25. Figure~\ref{fig:math500_experiment_comparison} further supports this picture from the training dynamics: Latent-GRPO steadily improves validation performance on Math500 while maintaining a much lower response length throughout training.

\begin{table*}[t]
\centering
\small
\setlength{\tabcolsep}{4.0pt}
\renewcommand{\arraystretch}{1.12}
\resizebox{\textwidth}{!}{%
\begin{tabular}{lcccccccccccc}
\toprule
& \multicolumn{2}{c}{Math500}
& \multicolumn{2}{c}{AIME24}
& \multicolumn{2}{c}{AIME25}
& \multicolumn{2}{c}{GPQA}
& \multicolumn{4}{c}{Average} \\
\cmidrule(lr){2-3}\cmidrule(lr){4-5}\cmidrule(lr){6-7}\cmidrule(lr){8-9}\cmidrule(lr){10-13}
Method
& Pass@1 $\uparrow$ & \#L $\downarrow$
& Pass@1 $\uparrow$ & \#L $\downarrow$
& Pass@1 $\uparrow$ & \#L $\downarrow$
& Pass@1 $\uparrow$ & \#L $\downarrow$
& Pass@1 $\uparrow$ & \#L $\downarrow$ & $\Delta$ Pass@1 $\uparrow$ & GRPO/\#L $\uparrow$ \\
\midrule
\multicolumn{13}{c}{\textbf{Qwen2.5-Math-7B}} \\
\cmidrule(lr){1-13}
\multicolumn{13}{c}{\textit{Explicit Reasoning}} \\
SFT
& 67.40 & 2662 & 13.39 & 6423 & 6.77 & 6550 & 32.07 & 3555 & 29.91 & 4798 & \tbdres & \tbdres \\
GRPO
& 75.15 & 2982 & 16.61 & 7500 & 19.79 & 6815 & \textbf{38.25} & 4568 & 37.45 & 5466 & +7.54 & 1.00$\times$ \\
\cmidrule(lr){1-13}
\multicolumn{13}{c}{\textit{Latent Reasoning (No Sampling)}} \\
Latent-SFT
& 64.85 & 1070 & 6.67 & 1651 & 6.61 & 1299 & 29.67 & 1214 & 26.95 & 1309 & \tbdres & \tbdres \\
Soft-GRPO
& 60.60 & 1133 & 6.20 & 1894 & 3.59 & 1511 & 27.12 & 1426 & 24.38 & 1491 & -2.57 & 3.67$\times$ \\
\rowcolor{latentgrpopurple}
Latent-GRPO
& \textbf{80.40} & 929 & \textbf{26.56} & 2548 & \textbf{23.23} & 1735 & 36.68 & 1384 & \textbf{41.72} & 1649 & \textbf{+14.77} & 3.31$\times$ \\
\bottomrule
\end{tabular}
}
\caption{Main results on the high-difficulty benchmark suite.}
\label{tab:high_difficulty_main_results}
\end{table*}

\subsection{Ablation Study}
Figures~\ref{fig:gsm8k_experiment_comparison} and~\ref{fig:math500_experiment_comparison} also provide the ablation results for the two key designs introduced in Section~\ref{sec:one_sided_noise_sampling} and Section~\ref{sec:first_token_selection}, namely \textbf{One-sided Noise Sampling} and \textbf{First Token Selection}. The most important factor is One-sided Noise Sampling. On both the low-difficulty GSM8K-Aug setting and the high-difficulty Math500 setting, removing this design causes the validation performance to drop sharply and eventually collapse, while the response length becomes much less controlled. This behavior is consistent with the motivation of Section~\ref{sec:one_sided_noise_sampling}: once the latent perturbation is no longer one-sided, the optimization direction can become misaligned with the trajectory advantage, so correct latent trajectories are no longer reinforced in a stable way.

By contrast, removing First Token Selection leads to a milder but still clear degradation. On the low-difficulty setting, the model can still achieve a noticeable performance increase in the early stage, showing that latent RL remains learnable without this design, but the improvement is consistently smaller than that of full Latent-GRPO and the training remains less stable. On the high-difficulty setting, the gap becomes much more pronounced, with severe instability appearing much earlier. This pattern also matches the role of Section~\ref{sec:first_token_selection}: when multiple correct latent paths are optimized together, mode averaging at the shared first latent token is especially damaging on harder reasoning problems. Overall, the ablation results show that One-sided Noise Sampling is the primary driver of training stability, while First Token Selection provides an additional improvement that is necessary for stronger and more reliable gains.

\subsection{Pass@k Performance of Latent-GRPO}
As described in Section~\ref{sec:latent_reasoning}, latent reasoning in the No Sampling mode is deterministic: given the same prompt, the model constructs the same latent tokens and therefore produces the same output across repeated runs. As a result, this setting is not suitable for measuring pass@$k$ on high-difficulty benchmarks. To evaluate the sampling capability of Latent-GRPO, we follow prior work~\citep{DBLP:journals/corr/abs-2508-03440}\footnote{For readers unfamiliar with this sampling formulation, we recommend reading ~\citet{DBLP:journals/corr/abs-2508-03440} in detail.} and inject Gumbel noise into the latent-token construction during inference. This perturbation enables diverse latent reasoning trajectories from the same prompt, making pass@$k$ evaluation meaningful while remaining within the latent reasoning framework.

\begin{figure}[t]
\centering
\includegraphics[width=\linewidth]{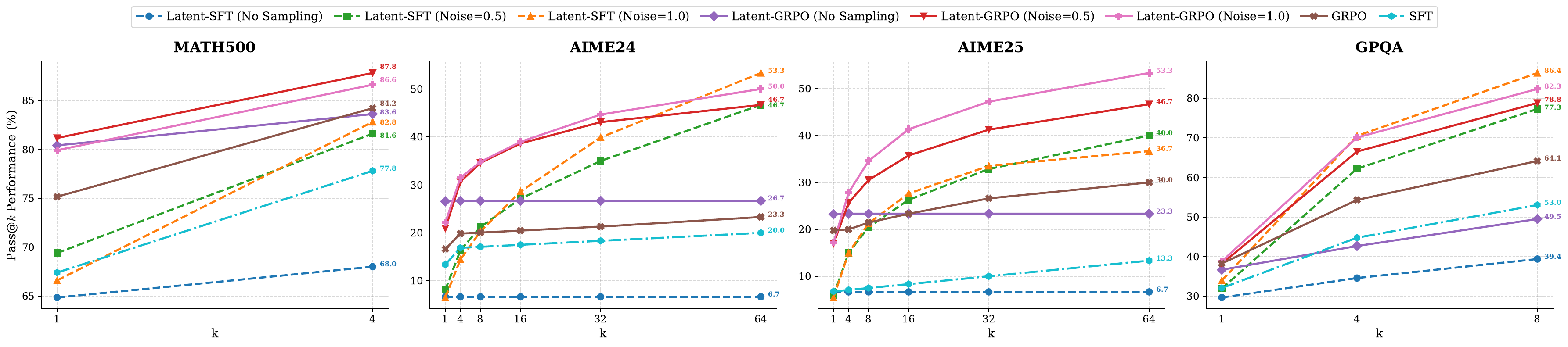}
\caption{Pass@$k$ performance of different models on the high-difficulty benchmarks.}
\label{fig:passk_comparison}
\end{figure}

Figure~\ref{fig:passk_comparison} shows that latent reasoning still exhibits strong pass@$k$ capability under Gumbel sampling. Compared with explicit GRPO, Latent-GRPO yields clearly stronger pass@$k$ curves on the mathematical benchmarks. On Math500, sampled Latent-GRPO reaches 87.8, surpassing the 84.2 achieved by GRPO. The gap is much larger on the harder AIME benchmarks: on AIME24, GRPO reaches 23.3 at pass@$64$, whereas Latent-GRPO reaches 46.7 and 50.0 under noise $=0.5$ and $1.0$; on AIME25, the corresponding improvement is from 30.0 to 46.7 and 53.3. Across different noise strengths, the Latent-GRPO curves are generally smooth and improve consistently as $k$ increases, indicating that latent reasoning can support effective stochastic exploration rather than collapsing under noisy decoding. The preferred noise level is somewhat dataset-dependent, but both noise $=0.5$ and noise $=1.0$ yield strong pass@$k$ performance overall. More broadly, increasing the noise strength generally encourages broader exploration over diverse latent reasoning paths, which is beneficial for discovering multiple valid trajectories in difficult reasoning problems.

At the same time, the figure also reveals an important trade-off on the hardest AIME benchmarks. After enabling Gumbel sampling, the Latent-GRPO model suffers a clear drop in pass@1 on both AIME24 and AIME25 compared with the no-sampling mode, but its pass@$k$ performance improves dramatically. On AIME24, the deterministic Latent-GRPO result is 26.7, whereas sampling raises the final pass@$64$ performance to 46.7 and 50.0 under noise $=0.5$ and $1.0$, respectively. On AIME25, the corresponding improvement is even larger, increasing from 23.3 to 46.7 and 53.3. A plausible explanation is that these benchmarks are sufficiently difficult that noise can perturb the model away from its single best trajectory, hurting pass@1, while still improving the diversity of sampled reasoning paths and therefore increasing the probability of obtaining a correct answer when multiple trials are allowed.

\section{Related Work}

\paragraph{LLMs and Explicit Reasoning.} Large language models have rapidly become a general foundation for a wide range of tasks ~\citep{DBLP:conf/iclr/DengWPDSC25, DBLP:journals/corr/abs-2601-06226}, including mathematical problem solving ~\citep{chen2024alphamath}, agent scenarios~\citep{DBLP:conf/nips/YaoYZS00N23, DBLP:conf/www/XuPSCC24,han2025see}, and other complex reasoning settings ~\citep{DBLP:conf/acl/CaoHGDXZ25, DBLP:journals/corr/abs-2601-05693, shi2026codehackerautomatedtestcase, Zhao_Min_Wu_Li_Sun_Cai_Wang_Chen_Penn_2026}. A central driver of this progress is \textit{explicit reasoning}, where the model produces discrete chains of thought to decompose a difficult problem into intermediate steps ~\citep{DBLP:conf/nips/Wei0SBIXCLZ22}. Subsequent test-time scaling studies further showed that performance can often be improved by allocating more reasoning tokens at inference time ~\citep{DBLP:conf/emnlp/MuennighoffYSLFHZLCH25, DBLP:journals/corr/abs-2412-16720}. In parallel, RL post-training has become an increasingly important way to strengthen explicit reasoning, especially in mathematical domains. In particular, GRPO shows that policy optimization over explicit reasoning trajectories can yield strong gains with a simple and effective group-relative objective ~\citep{DBLP:journals/corr/abs-2402-03300}. However, this paradigm remains inherently verbose: better performance is frequently obtained by generating longer and longer reasoning traces, which introduces substantial redundancy, latency, and computational cost. This limitation motivates latent reasoning, which shifts the intermediate reasoning process from explicit language into continuous representations in order to retain the benefits of reasoning while using far fewer reasoning tokens.

\paragraph{Training-free Latent Reasoning.} Training-free methods modify the inference process without updating model parameters. Soft Thinking constructs each latent token as a probability-weighted mixture of discrete token embeddings, showing that continuous concept tokens can improve reasoning to some extent while remaining interpretable ~\citep{DBLP:journals/corr/abs-2505-15778}. Follow-up analysis in ~\citep{DBLP:journals/corr/abs-2508-03440} argues that vanilla soft thinking is still dominated by the highest-probability token, which leads to a greedy feedback loop and motivates stochastic variants such as Gumbel-Softmax for stronger exploration. Parallel test-time scaling further extends inference-time exploration by combining stochastic latent sampling with a latent reward model for trajectory selection in continuous space ~\citep{DBLP:journals/corr/abs-2510-07745}. Overall, these methods show that inference-time stochasticity can make latent reasoning more exploratory, but their performance gains are generally modest and they do not establish the strong reasoning-length advantage targeted in our setting.

\paragraph{Training-based Latent Reasoning.} Training-based methods instead learn latent reasoning representations directly. Coconut first showed that feeding continuous hidden states back into the model can support latent reasoning and encode multiple alternative next steps during reasoning ~\citep{DBLP:journals/corr/abs-2412-06769}. CODI compresses explicit CoT into latent states via self-distillation and matches explicit CoT on GSM8K with substantial compression ~\citep{DBLP:conf/emnlp/ShenYZHDH25}. CoLaR dynamically compresses reasoning chains and further improves them with RL ~\citep{DBLP:journals/corr/abs-2505-16552}. Render-of-Thought moves latent reasoning into a visual space by rendering textual CoT as images, improving traceability while retaining compression ~\citep{DBLP:journals/corr/abs-2601-14750}. Latent-SFT develops a unified latent reasoning framework with latent vocabulary, latent chain construction, and stochastic optimization, showing that latent reasoning can outperform explicit SFT with shorter traces ~\citep{DBLP:journals/corr/abs-2510-15522}. More recent work explores RL and stochastic soft decoding more directly. Continuous CoT RL in ~\citep{DBLP:journals/corr/abs-2509-19170} introduces scalable reinforcement learning for continuous CoT, Multiplex Thinking performs token-wise branch-and-merge to optimize stochastic soft rollouts ~\citep{DBLP:journals/corr/abs-2601-08808}, and SofT-GRPO adapts GRPO to Gumbel-reparameterized soft thinking ~\citep{DBLP:journals/corr/abs-2511-06411}. Overall, training-based methods establish latent reasoning as a strong alternative to explicit CoT. By contrast, our work starts directly from a Latent-SFT model and studies how to achieve stable performance gains with only a small number of reasoning tokens, thereby obtaining dual benefits in both task performance and reasoning efficiency.

\section{Conclusion}
In this work, we study how to make reinforcement learning post-training effective for latent reasoning. We show that directly adapting GRPO to latent reasoning is fundamentally unstable because latent RL must simultaneously handle off-manifold exploration, optimization mismatch under noisy latent sampling, and harmful joint fitting over multiple correct latent paths. To address these issues, we propose Latent-GRPO, which combines invalid-sample advantage masking, one-sided noise sampling, and optimal correct-path first-token selection. Experiments on both low- and high-difficulty benchmarks show that Latent-GRPO achieves stable performance gains while substantially shortening reasoning chains relative to explicit GRPO, and our pass@$k$ and ablation results further verify its sampling capability and the importance of the key designs. Overall, our results suggest that stable RL post-training is a practical route toward efficient and high-performing latent reasoning.

\bibliography{references}
\bibliographystyle{unsrtnat}

\appendix

\section{Gradient Analysis of Exploration-Optimization Misalignment}
\label{app:gradient_misalignment}
We provide a gradient-level explanation for the Exploration-Optimization Misalignment discussed in Section~\ref{sec:one_sided_noise_sampling}. Consider one latent step and let $\mathcal{V}_t^{(K)}$ denote the top-$K$ candidate set used to construct the latent token, consistent with Section~\ref{sec:latent_reasoning}. During rollout, the perturbed score
\begin{equation}
q_i = g'_i = \log p_i(\theta_{\mathrm{old}}) + \xi_i
\end{equation}
is fixed for each $i\in\mathcal{V}_t^{(K)}$. During optimization, the current policy changes $\log p_i(\theta)$, so the standard Gumbel perturbation margin is
\begin{equation}
\Delta_i(\theta)=q_i-\log p_i(\theta).
\end{equation}
The latent-segment Gumbel log-density used by Soft-GRPO can be written as
\begin{equation}
\ell_G(\theta)
=
\sum_{i\in\mathcal{V}_t^{(K)}}
\left[
-\Delta_i(\theta)-\exp\!\left(-\Delta_i(\theta)\right)
\right].
\end{equation}
Differentiating with respect to the current log-probability of the $i$-th component gives
\begin{equation}
\frac{\partial \ell_G(\theta)}
{\partial \log p_i(\theta)}
=
1-\exp\!\left(-\Delta_i(\theta)\right).
\end{equation}
In the local region where PPO clipping is inactive and the importance ratio is close to one, the corresponding contribution to the GRPO update has gradient
\begin{equation}
\frac{\partial \mathcal{J}}
{\partial \log p_i(\theta)}
\approx
\hat{A}_j
\left[
1-\exp\!\left(-\Delta_i(\theta)\right)
\right].
\end{equation}
Therefore, for a positive-advantage trajectory, a component is reinforced only when $\Delta_i(\theta)>0$. If $\Delta_i(\theta)<0$, the same positive-advantage trajectory instead decreases $\log p_i(\theta)$. Conversely, for a negative-advantage trajectory, a negative margin can increase the component probability. Since the original Gumbel perturbation is two-sided, $\Delta_i(\theta_{\mathrm{old}})=\xi_i$ can be either positive or negative at the beginning of optimization. This is the gradient-level origin of Exploration-Optimization Misalignment.

One-sided Noise Sampling removes this sign ambiguity. In Latent-GRPO, the target is
\begin{equation}
g_i^*=\log p_i(\theta_{\mathrm{old}})+\xi_i^+,
\qquad
\xi_i^+>0,
\end{equation}
so at the beginning of optimization $\Delta_i(\theta_{\mathrm{old}})=\xi_i^+>0$. If repeated PPO updates later cross the fixed target and produce $\Delta_i(\theta)<0$, the conditional STE in Section~\ref{sec:one_sided_noise_sampling} flips the backward gradient of the margin. For the one-sided surrogate log-likelihood
\begin{equation}
\ell_G^+(\theta)
=
\sum_{i\in\mathcal{V}_t^{(K)}}
\left[
-\widetilde{\Delta}_i(\theta)
-\exp\!\left(-\widetilde{\Delta}_i(\theta)\right)
\right],
\end{equation}
the effective derivative becomes
\begin{equation}
\frac{\partial \ell_G^+(\theta)}
{\partial \log p_i(\theta)}
=
\begin{cases}
1-\exp\!\left(-\Delta_i(\theta)\right),
& \Delta_i(\theta)\ge 0, \\[1ex]
\exp\!\left(-\Delta_i(\theta)\right)-1,
& \Delta_i(\theta)<0.
\end{cases}
\end{equation}
This derivative is non-negative for all $\Delta_i(\theta)$ and is strictly positive whenever $\Delta_i(\theta)\ne 0$. Because the conditional STE modifies the backward pass, $\ell_G^+$ and $\pi_\theta^+$ should be understood as a surrogate latent likelihood with a custom backward rule rather than a mathematically exact probability density. Consequently, in the unclipped local regime,
\begin{equation}
\frac{\partial \mathcal{J}_{\mathrm{Latent\text{-}GRPO}}}
{\partial \log p_i(\theta)}
\approx
\hat{A}_j
\frac{\partial \ell_G^+(\theta)}
{\partial \log p_i(\theta)},
\end{equation}
so the direct component-wise score with respect to $\log p_i(\theta)$ is aligned with the trajectory advantage $\hat{A}_j$ rather than with the accidental sign of a two-sided Gumbel perturbation.

This conclusion is local to the coordinates $\log p_i(\theta)$ and should not be read as a guarantee that every selected raw logit increases individually. Under the full-softmax parameterization, let $z_l$ denote the raw logit of token $l$ and define
\begin{equation}
h_i=\frac{\partial \ell_G^+(\theta)}
{\partial \log p_i(\theta)},
\qquad
H=\sum_{i\in\mathcal{V}_t^{(K)}}h_i.
\end{equation}
Since
\begin{equation}
\frac{\partial \log p_i(\theta)}{\partial z_l}
=
\mathbf{1}[i=l]-p_l(\theta),
\end{equation}
we have
\begin{equation}
\frac{\partial \ell_G^+(\theta)}{\partial z_l}
=
\sum_{i\in\mathcal{V}_t^{(K)}}h_i\left(\mathbf{1}[i=l]-p_l(\theta)\right).
\end{equation}
Thus, for $l\in\mathcal{V}_t^{(K)}$,
\begin{equation}
\frac{\partial \ell_G^+(\theta)}{\partial z_l}
=
h_l-p_l(\theta)H,
\end{equation}
whose sign is not guaranteed to be positive for every selected component due to softmax competition. For $l\notin\mathcal{V}_t^{(K)}$,
\begin{equation}
\frac{\partial \ell_G^+(\theta)}{\partial z_l}
=
-p_l(\theta)H \le 0,
\end{equation}
so non-selected tokens receive an aggregate downward signal when $H\ge0$. Therefore, One-sided Noise Sampling aligns the direct component-wise score $\partial \ell_G^+/\partial \log p_i(\theta)$ with the trajectory advantage and induces an aligned aggregate learning signal under softmax normalization, but it does not strictly imply that every selected token logit increases individually.

\section{Implementation Details}
\label{app:implementation_details}
This appendix provides the detailed implementation settings used in our experiments.

\paragraph{Latent-SFT.}
We first describe the implementation details of the Latent-SFT initialization used in our experiments. Our implementation is built on the official Latent-SFT repository: \url{https://github.com/DJC-GO-SOLO/Latent-SFT}. To facilitate downstream Latent-GRPO training, we set $K=10$ in Section~\ref{sec:latent_reasoning}, so that each latent token is constructed from the top-10 explicit tokens. In practice, using a larger $K$ increases the risk of out-of-memory issues during Latent-GRPO training.

For the training setup, in Stage 1 we train 10 epochs per step with a learning rate of $5\times 10^{-5}$ and a compression ratio of 2. In Stage 2, the learning rate is set to $3\times 10^{-4}$ and training runs for 70 epochs. We further add the Gumbel training objective with Gumbel noise set to 1.0 and Gumbel temperature set to 1.0. More detailed training hyperparameters can be found in the corresponding scripts of the code repository.

\paragraph{Latent-GRPO.}
We then describe the implementation details of Latent-GRPO. For each setting, we select the best-performing Latent-SFT checkpoint as the initialization model for Latent-GRPO. The learning rate is fixed at $1\times 10^{-6}$ throughout training.

For LLaMA-3.2-1B-Instruct, we set \texttt{train\_batch\_size}=64, \texttt{ppo\_mini\_batch\_size}=16, the number of rollouts to 8, and \texttt{max\_response\_length}=128. For Qwen2.5-Math-7B, we use a different configuration with \texttt{train\_batch\_size}=32, \texttt{ppo\_mini\_batch\_size}=32, and \texttt{max\_response\_length}=4096. More detailed training hyperparameters can be found in the corresponding scripts of the code repository.

\paragraph{SFT.}
For completeness, we also summarize the implementation details of the explicit SFT baseline. We train the SFT models with LoRA, using a uniform rank of 64 and 10 training epochs for all settings. The learning rate is set to $5\times 10^{-5}$ for LLaMA-3.2-1B-Instruct and $2\times 10^{-5}$ for Qwen2.5-Math-7B.

\paragraph{GRPO.}
Finally, we summarize the implementation details of the explicit GRPO baseline. We use version 0.7.0 of \texttt{verl} and run the experiments with the official scripts. The learning rate is uniformly set to $1\times 10^{-6}$. For LLaMA-3.2-1B-Instruct, we set \texttt{max\_response\_length}=512. For Qwen2.5-Math-7B, we set \texttt{max\_response\_length}=8192.

All experiments are conducted on 8 A100 GPUs.


\end{document}